\documentclass[conference,compsoc]{IEEEtran}
\usepackage{amsmath}
\usepackage{amsthm}
\usepackage{amssymb}
\usepackage{amsfonts}
\usepackage{xcolor}
\usepackage{graphicx} 
\usepackage{subcaption}
\usepackage[hidelinks]{hyperref}

\usepackage{url}

\newtheorem{proposition}{Proposition}
\usepackage{caption}
\newtheorem{theorem}{Theorem} 

\usepackage{subcaption}

\usepackage{pgfplots}

\ifCLASSOPTIONcompsoc
  \usepackage[nocompress]{cite}
\else
  \usepackage{cite}
\fi
\ifCLASSINFOpdf
\else
\fi
\begin{document}
%


\title{Impact of Data Duplication on Deep Neural Network-Based Image Classifiers: Robust vs. Standard Models}


\author{\IEEEauthorblockN{Alireza Aghabagherloo\IEEEauthorrefmark{1}\IEEEauthorrefmark{4}, Aydin Abadi\IEEEauthorrefmark{7}, Sumanta Sarkar\IEEEauthorrefmark{6}, Vishnu Asutosh Dasu\IEEEauthorrefmark{5} and Bart Preneel\IEEEauthorrefmark{1}\IEEEauthorrefmark{4}} 
    \IEEEauthorblockA{\IEEEauthorrefmark{1} COSIC, Department of Electrical Engineering, KU Leuven, 3001 Leuven, Belgium \\
    \IEEEauthorrefmark{4}{\{Alireza.aghabagherloo, bart.preneel\}@esat.kuleuven.be}}
    \IEEEauthorblockA{\IEEEauthorrefmark{7} Newcastle University, aydin.abadi@ncl.ac.uk}
    \IEEEauthorblockA{\IEEEauthorrefmark{6} University of Essex, Sumanta.Sarkar@essex.ac.uk}
    \IEEEauthorblockA{\IEEEauthorrefmark{5} Pennsylvania State University, vdasu@psu.edu}

}


%


\maketitle

\begin{abstract}

The accuracy and robustness of machine learning models against adversarial attacks are significantly influenced by factors such as training data quality, model architecture, the training process, and the deployment environment. In recent years, duplicated data in training sets, especially in language models, has attracted considerable attention. It has been shown that deduplication enhances both training performance and model accuracy in language models. 
While the importance of data quality in training image classifier Deep Neural Networks (DNNs) is widely recognized, the impact of duplicated images in the training set on model generalization and performance has received little attention. 

In this paper, we address this gap and provide a comprehensive study on the effect of duplicates in image classification. Our analysis indicates that the presence of duplicated images in the training set not only negatively affects the efficiency of model training but also may result in lower accuracy of the image classifier. This negative impact of duplication on accuracy is particularly evident when duplicated data is non-uniform across classes or when duplication, whether uniform or non-uniform, occurs in the training set of an adversarially trained model. Even when duplicated samples are selected in a uniform way, increasing the amount of duplication does not lead to a significant improvement in accuracy.

\end{abstract}

\footnote{This paper has been published in IEEE. The final version is available at:https://doi.org/10.1109/SPW67851.2025.00023}

%
\IEEEpeerreviewmaketitle


\section{Introduction}
Machine learning (ML) is a data-driven process that heavily relies on training datasets to develop efficient and robust models. During training, an ML model learns patterns and relationships within the dataset. High-quality data contributes to more accurate and reliable models, whereas poor-quality data can introduce errors or biases \cite{budach2022effects}. Numerous studies \cite{budach2022effects, fenza2021dataset, esposito2024validate, xiong2024data} indicate that vulnerabilities often stem from the dataset itself rather than the model. These works emphasize that issues such as poor data quality, bias, sparsity, or misrepresentation can significantly affect model performance and robustness, leading to vulnerabilities independent of the model's architecture.

Thus, conducting a thorough analysis and evaluation of dataset quality is crucial for ensuring the development of effective data \textit{preprocessing} methods. Preprocessing involves cleaning raw data through various steps, including removing redundancies and outliers. The recently proposed DeepSeek language model~\cite{deepseek}, which has garnered significant attention, also highlights the importance of high-quality data in building efficient models.
One concern that may hinder the efficiency of a language model is the presence of duplicate entries in text datasets, as discussed in \cite{lee-etal-2022-deduplicating}, which emphasizes the importance of removing duplicates before training. 

Similarly, duplicate images frequently appear in real-world datasets. Consequently, a similar conclusion can be drawn that duplicates may affect ML models for image processing, particularly in image classification tasks.
However, the impact of duplication on image classification has received limited attention. Previous studies on image classification have primarily examined the effect of training set images being duplicated on the test sets \cite{barz2020we, li2021ce}. 

We hypothesize that the existence of duplicates within the training set in image classifier DNNs will also prohibit general learning and will negatively affect efficiency, accuracy, and robustness against attacks like adversarial attacks\cite{aghabagherloo2023robust, Ilyas2019Madry}. However, existing literature does not examine the impact of duplicates in training data, which leaves their effects unexplored.

In this work, we address these gaps in the literature by conducting a comprehensive study on the impact of duplicates in image classification.  We examine the impact of varying levels of duplication on model generalization and training efficiency by considering both uniform and non-uniform duplication scenarios. We also evaluate the impact of repetitive data within the dataset of an adversarially trained model~\cite{Tramer2019}, shedding light on its influence on robustness and accuracy.

\subsection{Our Contribution}
\label{Our contribution}

The previous works studying the effect of repetitive data within datasets indicated that duplication can increase accuracy and result in better generalization while being susceptible to memorization. In this paper, our primary goal is to obtain a complete understanding of the impact of image duplication in the training set on generalization. We show that duplication in image classifier datasets does not always yield better generalization and can negatively affect it, especially in non-uniform duplications and adversarial settings. 

Our primary goal is to present a comprehensive analysis of how duplicates affect the accuracy of image classification.
We will first conduct a theoretical analysis of the impact of duplication in different scenarios. Assuming data is selected from two Gaussian distributions, we will evaluate the impact of adding duplicate data. Selecting both non-uniform and uniform duplication with regard to duplicated data point labels, our analysis shows that non-uniform duplication leads to biased decision boundaries and reduces model generalization to unseen data. Our empirical results on the CIFAR-10 dataset, when randomly selected images have been duplicated, show no considerable accuracy improvement. A similar experiment on the adversarially trained model on the CIFAR-10 dataset shows an accuracy drop when adding duplicated images. The implementations are publicly available \cite{sourcecode}. 
The main contributions of this paper include:

\begin{itemize}
    \item A theoretical analysis on the effect of duplication in generalization in standard and adversarially trained models will be given in Section~\ref{Theoretical preliminaries}.
    \item An experiment-based analysis of the impact of repetitive data, focusing on data collected from two Gaussian distributions, examining the effects of duplication when the duplicated data is either uniform or non-uniform, will be given in Section~\ref{Theoretical Analysis of the Impact of Repetitive Data on Different Scenarios1}.
    \item An experimental-based analysis of the impact of duplication when data is selected from the CIFAR-10 dataset in both standard models trained on CIFAR-10 and an adversarially trained model on CIFAR-10 will be given in Section~\ref{Our results}.

\end{itemize}

The paper is structured as follows: Section 2 reviews the related work on the impact of duplicates on the generalization error. Section 3 presents the theoretical perspective on how data duplication affects generalization error. Section 4 provides an experimental-based analysis of the impact of repetitive data. Section 5 concludes the paper.

\section{Related Work}

Data duplication, which often occurs in the real world, poses a critical concern in training datasets intended for the training of an ML model\cite{abadi2024privacy}. As mentioned earlier, despite its importance, this issue has received relatively less attention in image classification than in other areas, such as Large Language Models (LLMs). In the context of LLMs, much research has been conducted on the impact of duplicate data  \cite{abadi2024privacy, carlini2024memorization, lee2022deduplicating}. Lee et al.\cite{lee2022deduplicating} show that duplication in training data of language models results in the memorization of duplicated data. Carlini et al. argue that memorization has adverse effects on the privacy and fairness of LLMs. Memorization makes LLMs vulnerable to membership inference \cite {shokri2017membership} and data extraction \cite{carlini2021extracting} attacks. Abadi et al. \cite{abadi2024privacy} address the duplication issue in Federated Learning (FL) based LLMs across clients, introducing a privacy-preserving data de-duplication mechanism. This body of work emphasizes how duplicates can skew model training and performance in LLMs, leading to overfitting and reduced generalization. In LLMs, it has also been shown that deduplicated datasets reduce the memorization frequency and improve generalization  \cite{lee2022deduplicating}. 

In contrast, the impact of duplication in image classification has not been systematically studied. Only a limited set of studies have focused on this issue.  For example, Barz and Denzler \cite{barz2020we} highlight a considerable volume of exact duplicates and near-duplicates between the test and training sets in CIFAR-10 and CIFAR-100 \cite{krizhevsky2009learning} datasets.

Their studies show that when they tested on ciFAIR (the duplicate-free version of CIFAR), the models experienced a substantial drop in classification accuracy; in particular, in CIFAR-10, there is a 9\%–14\% drop in accuracy when they removed repeated images between test set and train set. In contrast, with LLM,  where memorization resulting from data duplication harms generalization~\cite{abadi2024privacy,lee2022deduplicating}, limited research on image classifier DNNs suggests that duplication may actually enhance generalization~\cite{barz2020we,feldman2020memorize}. 
%

Similarly,~\cite{abdullah2021generalization} explores whether memorization resulting from duplication of training data in a test set is necessary for achieving generalization in models, especially in the deep learning context. The main limitation of their observation is that they mainly examine the effect of the duplication of train sets in the test set rather than evaluating the impact of having duplicated data when training. A limitation that we aim to address in this work. 

Another study by Li et al. \cite{li2021ce} introduces a framework called CE-Dedup, which evaluates the impact of near-duplicate images on Convolutional Neural Networks (CNN) training performance. The authors propose a hash-based image deduplication method to balance the trade-off between dataset size and model accuracy. Their experiments demonstrate that removing redundant images can reduce dataset size by up to 75\% with an accuracy loss, leading to more efficient training. However, this work mainly focuses on a deduplication method rather than providing a detailed analysis of the effects of duplication on model performance.

In this paper, we present a comprehensive analysis of the impact of data duplication across various scenarios, demonstrating that while duplication in image classifiers always sacrifices efficiency, it does not necessarily lead to improved accuracy. More focus is on the effect of duplication in training sets on direct generalization.
Our experiments on the CIFAR-10 dataset reveal that exceeding a certain threshold of duplicated randomly selected images leads to overfitting. In adversarially trained models, duplication further degrades accuracy. This shows that, above the mentioned threshold in standard training or an adversarial training setting, deduplication increases efficiency (as discussed in~\cite{li2021ce}) and results in better generalization. 

\section{Theoretical Perspective on the Impact of Duplicates on Generalization Error}
\label{Theoretical preliminaries}

\subsection{Generalization Error}

Generalization refers to a model's ability to perform well on unseen data, balancing between bias and variance. The generalization error can be decomposed into three components: bias, variance, and irreducible error. Geman et al.~\cite{geman1992biasvariance} decomposed the Mean Squared Error (MSE) loss function in terms of a prediction model's bias and variance, as stated in Theorem \ref{theorem::gen-error}~\cite{geman1992biasvariance}.


\begin{theorem} \label{theorem::gen-error} For a prediction model \(\hat{f}(x)\) trained on a dataset \(D\) to estimate the target function \(f(x)\) using an MSE loss function, the bias-variance trade-off is given by:


\begin{align}
\mathbb{E}_{x,D,\gamma} \left[ (y - \hat{f}(x))^2 \right] 
&= \mathbb{E}_{x,D} \left[ \left( \mathbb{E}_D[\hat{f}] - f(x) \right)^2 \right] \nonumber \\
&\quad + \mathbb{E}_{x,D} \left[ \left( \hat{f} - \mathbb{E}_D[\hat{f}] \right)^2 \right] \nonumber \\
&\quad + \sigma^2_\gamma.
\end{align}

\end{theorem}

This can be simplified as:

\begin{align}
\label{(1)}
e = \text{Bias}^2[\hat{f}] + \text{Var}[\hat{f}] + \text{Irreducible error}.
\end{align}

In this framework, bias decreases as the model complexity increases, allowing it to learn more patterns. However, this often leads to higher variance (overfitting), where the model becomes sensitive to the specific training data, potentially reducing generalization performance. In other words, the risk of overfitting typically decreases as the number of samples increases or the complexity of the model is reduced. However, reducing model complexity or increasing the number of samples can lead to higher bias. The trade-off between bias and variance is critical for improving model generalization without overfitting, particularly in repeated patterns or data duplication scenarios. 


\begin{proposition}\label{proposition:1}
    In a typical scenario, increasing the amount of data generally reduces variance. However, in the case of a duplicated dataset, the variance of the classes that do not have duplicated data in the training set may increase, while the variance of the classes with more duplicated data progressively decreases. Regarding bias, duplication amplifies bias in favor of the duplicated class. This dynamic underscores that generalization can either improve or deteriorate depending on the balance between variance and bias in the context of duplication.
\end{proposition}

In Section~\ref{Theoretical Analysis of the Impact of Repetitive Data on Different Scenarios}, we validate the bias-variance trade-off in a model trained on data with duplication established in Proposition~\ref{proposition:1}.

\subsection{Generalization Error in Adversarial Settings}
\label{Generalization-Error-in-Adversarial-Settings}
Typically, models exhibit susceptibility to adversarial attacks~\cite{pitropakis2019taxonomy,aghabagherloo2023robust,Ilyas2019Madry}, with adversarial training serving as a common defense mechanism. It involves augmenting the training dataset with adversarial examples (perturbed inputs) so the model learns to recognize and correctly classify them. This improves the model's resilience to adversarial perturbations by making the model robust to a wider range of potential attacks. As we will discuss, duplicated data might benefit adversaries. Thus, it is vital to analyze the effect of data duplication in the adversarial settings. 
%

Assume \( \bar{f}(x) = \mathbb{E}_D[\hat{f}(x)] \) and the target function is \( f(x) \). The bias-variance trade-off for the mean squared error (MSE) loss function becomes more complex when noise \( \gamma \) and adversarial perturbations \( \beta(x) \), generated by an adversarial algorithm, are introduced.
In adversarial machine learning, generalization becomes more challenging due to adversarial perturbations. These perturbations aim to exploit vulnerabilities in the model, often leading to a trade-off between robustness and generalization.

According to \cite{aboutalebi2020vulnerability}, the vulnerability of ML models under adversarial attacks can be attributed to the bias-variance trade-off. The relationship between generalization and adversarial vulnerability can be analyzed using the bias-variance decomposition in the context of adversarial settings, as stated in Theorem~\ref{theorem::adv}\cite{aboutalebi2020vulnerability}.



\begin{theorem}\label{theorem::adv}
Assume \( \bar{f}(x) = \mathbb{E}_D[\hat{f}(x)] \) represents the expected model prediction over the dataset \( D \), and \( f(x) \) is the true target function. When adversarial perturbations \( \beta(x) \) are introduced, the MSE loss function for a model \( \hat{f}(x) \) trained on \( D \) with noise \( \gamma \) is given by:

\begin{align}
    \mathbb{E}_{x,D,\gamma} \left[ (y - \hat{f}(x + \beta(x)))^2 \right] &\approx\\ 
    \mathbb{E}_{x,D} \left[ (f(x) - \bar{f}(x) - c_x)^2 \right] \nonumber 
     &+ \text{Var}[\gamma] + \text{Var}[\hat{f}] + \mathbb{E}_{x,D}[c'_x]
\end{align}

Where:
\[
    c_x = \nabla \bar{f}(x)^T \beta(x),
\]
and
\[
    c'_x = 2 \left( \hat{f}(x) - \bar{f}(x) \right) \left( \nabla \hat{f}(x) - \nabla \bar{f}(x) \right)^T \beta(x).
\]
\end{theorem}

In this equation, \( c_x \) represents the interaction between the model’s gradient and the adversarial perturbation, while \( c'_x \) captures the variability in the model’s predictions under adversarial conditions. These terms illustrate how adversarial perturbations can increase both bias and variance, thus impacting generalization in adversarial settings. This equation highlights that in an adversarial-trained model, beyond the typical trade-off between bias and variance, there is an additional term \( \mathbb{E}_{x,D}[c'_x] \). 
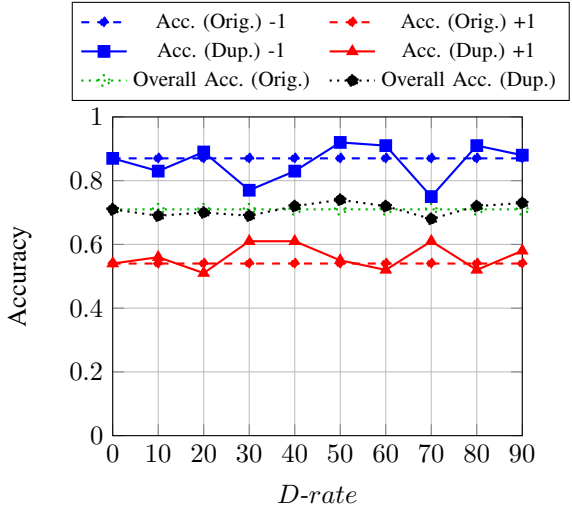
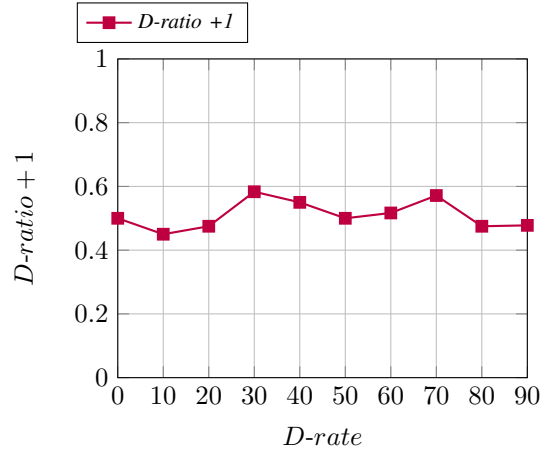
\begin{figure*}[ht!]
    \centering
    \begin{subfigure}[b]{0.45\textwidth}
        \centering
        \begin{tikzpicture}
            \begin{axis}[
                width=7cm,
                height=5.8cm,
                xlabel={$D\text{-}rate$},
                ylabel={Accuracy},
                grid=both,
                xmin=0, xmax=90,
                ymin=0, ymax=1,
                xtick={0,10,20,30,40,50,60,70,80,90},
                legend style={
                    at={(axis description cs:-.1,1.05)},
                    anchor=south west,
                    draw=black,
                    fill=white,
                    legend columns=2,
                    font=\footnotesize,
                    /tikz/every even column/.append style={column sep=0.5em},
                },
            ]
            \addplot[color=blue, dashed, mark=diamond*, thick] coordinates {
                (0,0.87) (10,0.87) (20,0.87) (30,0.87) (40,0.87) (50,0.87)
                (60,0.87) (70,0.87) (80,0.87) (90,0.87)
            };
            \addlegendentry{\footnotesize Acc. (Orig.) -1}
            
            \addplot[color=red, dashed, mark=diamond*, thick] coordinates {
                (0,0.54) (10,0.54) (20,0.54) (30,0.54) (40,0.54) (50,0.54)
                (60,0.54) (70,0.54) (80,0.54) (90,0.54)
            };
            \addlegendentry{\footnotesize Acc. (Orig.) +1}
            
            \addplot[color=blue, solid, mark=square*, thick] coordinates {
                (0,0.87) (10,0.83) (20,0.89) (30,0.77) (40,0.83) (50,0.92)
                (60,0.91) (70,0.75) (80,0.91) (90,0.88)
            };
            \addlegendentry{\footnotesize Acc. (Dup.) -1}
            
            \addplot[color=red, solid, mark=triangle*, thick] coordinates {
                (0,0.54) (10,0.56) (20,0.51) (30,0.61) (40,0.61) (50,0.55)
                (60,0.52) (70,0.61) (80,0.52) (90,0.58)
            };
            \addlegendentry{\footnotesize Acc. (Dup.) +1}
            
            \addplot[color=green!70!black, dotted, mark=o, thick] coordinates {
                (0,0.71) (10,0.71) (20,0.71) (30,0.71) (40,0.71) (50,0.71)
                (60,0.71) (70,0.71) (80,0.71) (90,0.71)
            };
            \addlegendentry{\footnotesize Overall Acc. (Orig.)}
            
            \addplot[color=black, dotted, mark=*, thick] coordinates {
                (0,0.71) (10,0.69) (20,0.70) (30,0.69) (40,0.72) (50,0.74)
                (60,0.72) (70,0.68) (80,0.72) (90,0.73)
            };
            \addlegendentry{\footnotesize Overall Acc. (Dup.)}
            \end{axis}
        \end{tikzpicture}
        \caption{Effect of duplication rate on accuracy results.}
        \label{fig:balanced_results}
    \end{subfigure}
    \hfill
    \begin{subfigure}[b]{0.45\textwidth}
        \centering
        \begin{tikzpicture}
            \begin{axis}[
                width=7cm,
                height=5.8cm,
                xlabel={$D\text{-}rate$},
                ylabel={$D\text{-}ratio+1$},
                grid=both,
                xmin=0, xmax=90,
                ymin=0.0, ymax=1,
                xtick={0,10,20,30,40,50,60,70,80,90},
                legend style={
                    at={(axis description cs:-.1,1.05)},
                    anchor=south west,
                    draw=black,
                    fill=white,
                    font=\footnotesize,
                },
            ]
            \addplot[color=purple, solid, mark=square*, thick] coordinates {
                (0,0.50) (10,0.45) (20,0.475) (30,0.5833) (40,0.55) 
                (50,0.50) (60,0.5167) (70,0.5714) (80,0.475) (90,0.4778)
            };
            \addlegendentry{\footnotesize \textit{D-ratio +1} 
}
            \end{axis}
        \end{tikzpicture}
        \caption{\textit{D-ratio +1} vs. \textit{D-rate}.}
        \label{fig:duplication_percentage}
    \end{subfigure}
    \caption{Comparison of accuracy results (left subfigure) and the percentage of duplications from class +1 to the total number of duplications (right subfigure, denoted as \textit{Duplication Ratio for +1 (D-ratio +1)}) with varying levels of data duplication (denoted as \textit{Duplication Rate (D-rate)}) in a uniform duplication setting.  
    \label{fig:repetition_effect_G1}
}
\end{figure*}

\begin{proposition} \label{proposition:2}
    
The impact of the additional term \( \mathbb{E}_{x,D}[c'_x] \) becomes significant when dealing with duplicated data. Specifically, duplicated data exacerbates the variability introduced by adversarial perturbations, leading to a compounding effect on bias and variance. This makes a model overly sensitive to specific perturbations within the duplicated data, reducing its ability to generalize effectively in adversarial settings to clear data. 
\end{proposition}

Section~\ref{Our results} discusses our experimental results on the accuracy and robustness of a model in an adversarial setting when we have duplicated data in the training set, validating the assertion in Proposition~\ref{proposition:2}.

\section{Experimental Analysis of the Impact of Duplication in the Training Set}
\label{Theoretical Analysis of the Impact of Repetitive Data on Different Scenarios}
In this section, we validate the theoretical analysis presented in the previous section using experimental results. First, we conduct theoretical experiments using data sampled from two Gaussian distributions. Then, we perform experiments on the CIFAR-10 dataset~\cite{Krizhevsky2009} in both standard and adversarial settings. Notably, in our paper, the standard setting refers to training a model, known as the standard model, without any adversarial training or robustification. In contrast, the adversarial setting denotes training a model that has been made robust using adversarial training~\cite{aghabagherloo2023robust}.

In our paper \textit{duplication} refers to the process of repeating and adding certain data points to the training set.  In our experiments, we selected duplications using two types of methods. The first method, \textit{uniform duplication}, includes uniformly selecting data from the dataset at random and adding them to the dataset. The second method, \textit{non-uniform (biased) duplication}, selects data points from one class more favorably than other classes.    

\subsection{Generalization Error with Gaussian-Distributed Data}
\label{Theoretical Analysis of the Impact of Repetitive Data on Different Scenarios1}

This section analyzes the effects of repetitive data across various scenarios, using two-dimensional Gaussian data generated with class means of $\left[0, 0\right]$ and $\left[1, 1\right]$, a shared covariance matrix of 
$\begin{bmatrix} 1 & 0.5 \\ 0.5 & 1 \end{bmatrix}$, and 100 samples per class. Then, duplicated data with varying \textit{Duplication Rates (D-rate)}, ranging from 10\% to 90\% of the dataset, have been added to the dataset to analyze the effect of duplication. Additionally, the Support Vector Machine (SVM) with a Radial Basis Function (RBF) kernel is used to compute decision boundaries in the experiments. The definitions of the RBF kernel, the optimization problem, and the mathematical details of the probability density functions for the two distributions are presented in Appendix \ref{sec::Theoretical-Model}. 

The impact of duplication on the decision boundary and generalization appears to be influenced by two key factors. The first factor is the randomly selected duplicates' pattern and proximity to the decision boundary. The second factor is the proportion of duplicated data points from a single class relative to the total dataset size, referred to as \textit{D-ratio +1}, which will be examined in Section~\ref{Balanced Duplication} and Section~\ref{Unbalanced Duplication}.
%


\subsubsection{Uniform Duplication}

\label{Balanced Duplication}


When uniform duplicates are added, the SVM's objective function experiences an increase in the effective density of data points near the mean vectors $\boldsymbol{\mu}_1$ and $\boldsymbol{\mu}_2$. This can cause the decision boundary to shift slightly towards regions with more duplicates. As the data is selected randomly (not favoring any specific class), this can, by chance, cause \textit{D-ratio +1} to deviate from 50\% but will generally remain close to it.
\begin{figure*}[ht!]
    \centering
    \begin{subfigure}[b]{0.45\textwidth}
        \centering
        \begin{tikzpicture}
            \begin{axis}[
                width=7cm,
                height=5.8cm,
                xlabel={$D\text{-}rate$},
                ylabel={Accuracy},
                grid=both,
                xmin=0, xmax=90,
                ymin=0, ymax=1.0,
                xtick={0,10,20,30,40,50,60,70,80,90},
                legend style={
                    at={(axis description cs:-.1,1.05)},
                    anchor=south west,
                    draw=black,
                    fill=white,
                    legend columns=2,
                    font=\footnotesize,
                    /tikz/every even column/.append style={column sep=0.5em},
                },
            ]
            \addplot[color=blue, dashed, mark=diamond*, thick] coordinates {
                (0,0.85) (10,0.85) (20,0.85) (30,0.85) (40,0.85) (50,0.85)
                (60,0.85) (70,0.85) (80,0.85) (90,0.85)
            };
            \addlegendentry{\footnotesize Acc. (Orig.) -1}
            
            \addplot[color=red, dashed, mark=diamond*, thick] coordinates {
                (0,0.67) (10,0.67) (20,0.67) (30,0.67) (40,0.67) (50,0.67)
                (60,0.67) (70,0.67) (80,0.67) (90,0.67)
            };
            \addlegendentry{\footnotesize Acc. (Orig.) +1}
            
            \addplot[color=blue, solid, mark=square*, thick] coordinates {
                (0,0.85) (10,0.75) (20,0.61) (30,0.56) (40,0.61) (50,0.66)
                (60,0.40) (70,0.38) (80,0.40) (90,0.42)
            };
            \addlegendentry{\footnotesize Acc. (Dup.) -1}
            
            \addplot[color=red, solid, mark=triangle*, thick] coordinates {
                (0,0.67) (10,0.78) (20,0.80) (30,0.81) (40,0.83) (50,0.76)
                (60,0.91) (70,0.96) (80,0.95) (90,0.94)
            };
            \addlegendentry{\footnotesize Acc. (Dup.) +1}
            
            \addplot[color=green!70!black, dotted, mark=o, thick] coordinates {
                (0,0.76) (10,0.76) (20,0.76) (30,0.76) (40,0.76) (50,0.76)
                (60,0.76) (70,0.76) (80,0.76) (90,0.76)
            };
            \addlegendentry{\footnotesize Overall Acc. (Orig.)}
            
            \addplot[color=black, dotted, mark=*, thick] coordinates {
                (0,0.76) (10,0.76) (20,0.71) (30,0.69) (40,0.72) (50,0.71)
                (60,0.66) (70,0.67) (80,0.67) (90,0.68)
            };
            \addlegendentry{\footnotesize Overall Acc. (Dup.)}
            \end{axis}
        \end{tikzpicture}
        \caption{Effect of duplicates on accuracy results.}
        \label{fig:balanced_results_G2}
    \end{subfigure}
    \hfill
    \begin{subfigure}[b]{0.45\textwidth}
        \centering
        \begin{tikzpicture}
            \begin{axis}[
                width=7cm,
                height=5.8cm,
                xlabel={$D\text{-}rate$},
                ylabel={$D\text{-}ratio +1$},
                grid=both,
                xmin=0, xmax=90,
                ymin=0, ymax=1,
                xtick={0,10,20,30,40,50,60,70,80,90},
                legend style={
                    at={(axis description cs:-.1,1.05)},
                    anchor=south west,
                    draw=black,
                    fill=white,
                    font=\footnotesize,
                },
            ]
            \addplot[color=purple, solid, mark=square*, thick] coordinates {
                (0,0.7) (10,0.725) (20,0.8625) (30,0.8667) (40,0.80625) 
                (50,0.745) (60,0.8) (70,0.7679) (80,0.7938) (90,0.7694)
            };
            \addlegendentry{\footnotesize \textit{D-ratio +1}}
            \end{axis}
        \end{tikzpicture}
        \caption{\textit{D-ratio +1} vs. \textit{D\text{-}rate}.}
        \label{fig:duplication_percentage_G2}
    \end{subfigure}
    \caption{Comparison of accuracy results (left subfigure) and the percentage of duplications from class +1 to the whole number of duplications (right subfigure) with varying levels of data duplication in a non-uniform setting.}
    \label{fig:repetition_effect_G2}
\end{figure*}
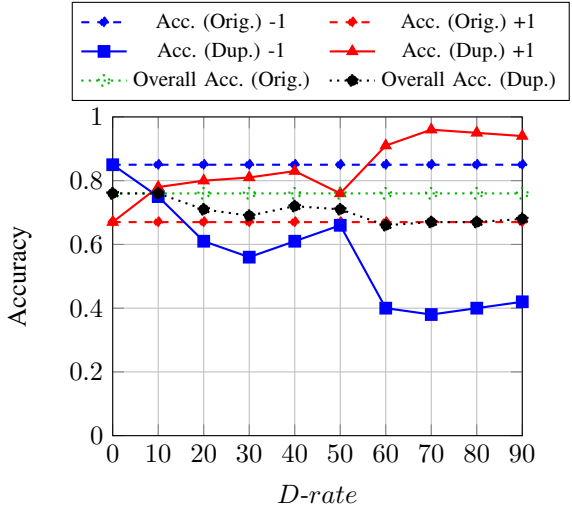
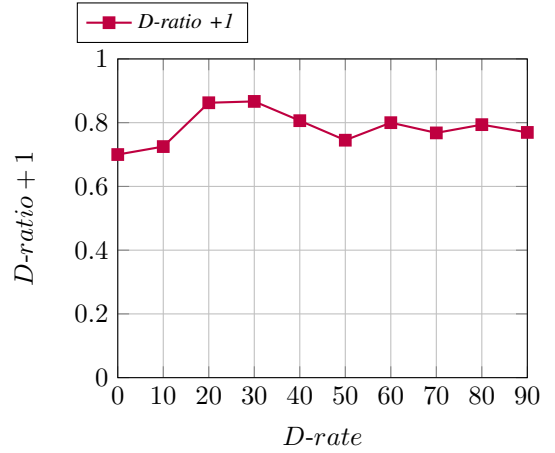
The detailed results are shown in Figure~\ref{fig:repetition_effect_G1}. We refer to the original dataset as data without any duplicates. \textbf{Acc. (Orig.) -1} and \textbf{Acc. (Orig.) +1} refer to the accuracy of class -1 and class +1 in the original dataset respectively, which means the rate of correctly identifying elements from class -1 and class +1, respectively. \textbf{Acc. (Dup.) -1} and \textbf{Acc. (Dup.) +1} represent the accuracy for these classes after duplicates are added to the original dataset. \textbf{Overall Acc. (Orig.)} and \textbf{Overall Acc. (Dup.)} denote the overall accuracy for the original and dataset with duplication, respectively.

A comparison of Figure~\ref{fig:balanced_results} and Figure~\ref{fig:duplication_percentage} reveals a correlation between \textit{D-ratio +1} and accuracy. Initially, class +1 has an accuracy of 87\%, while class -1 has 54\%. When \textit{D-ratio +1} is exactly 50\% (e.g., at \textit{D-rate} = 50), duplication increases accuracy, resulting in a 3\% increase. However, when \textit{D-ratio +1} deviates from 50\% (e.g., at \textit{D-rate} = 30), it benefits one class while harming the other, ultimately reducing overall accuracy. Specifically, at \textit{D-rate} = 30, \textit{D-ratio +1} is 58\%, increasing class +1 accuracy from 54\% to 61\% but decreasing class -1 accuracy from 87\% to 77\%, leading to a 2\% drop in total accuracy. The other important observation is that the effect of uniform duplicates on the accuracy of individual classes and the overall dataset remains within approximately the range of up to 10\% around the original accuracies.

This demonstrates that deviation from uniform duplication can introduce class bias, as described in Equation~\ref{(1)} and Proposition~\ref{proposition:1}, ultimately degrading accuracy. Another key observation from Figure~\ref{fig:balanced_results} is that increasing the number of duplications does not necessarily enhance generalization.

Therefore, uniform duplication generally has a positive effect on accuracy, except when there is a deviation from uniformity. To better understand how such deviations impact accuracy, we conducted an experiment where the number of duplications was biased toward a specific class.

\subsubsection{Non-uniform Duplication}
\label{Unbalanced Duplication}

Here, a non-uniform duplication strategy is applied by assigning different selection probabilities to the classes. Samples with label \(1\) are selected with a probability of 0.7, while those with label \(-1\) are selected with a probability of 0.3. As shown in Figure~\ref{fig:repetition_effect_G2} when duplicates are added from one class, say $\mathcal{C}_1$, the optimization problem's constraints are more heavily influenced by $\mathcal{C}_1$, resulting in a decision boundary that skews towards $\mathcal{C}_2$. This non-uniformity in the distribution of duplication increases the classifier's bias towards $\mathcal{C}_1$ and negatively impacts generalization.  

This figure clearly illustrates how non-uniform duplication introduces bias, leading to an increase in accuracy for one class while decreasing accuracy for the other. For instance, at \textit{D-rate} = 40, the original accuracy of Class -1 is 0.85, while Class +1 starts at 0.67. After duplication, the accuracy of Class -1 drops to 0.61, whereas the accuracy of Class +1 increases to 0.83. Consequently, the overall accuracy decreases from 0.76 to 0.72. Another key observation is that, in this setting, increasing the number of duplications further exacerbates bias and leads to a decrease in overall accuracy (Figure~\ref{fig:balanced_results_G2}).

\subsection{Generalization Error with CIFAR-10 Dataset}
\label{Our results}

In this section, we present an experimental-based analysis of the impact of data duplication when training models on the CIFAR-10 dataset. Our primary focus is to systematically analyze how randomly selected repetitive data influences classification accuracy in both standard models and an adversarially trained model, which is trained under Projected Gradient Descent (PGD) attacks constrained by an $\ell_2$ norm with $\epsilon = 0.5$. We introduce varying levels of duplicated data during training and evaluate model performance under each duplication setting.

\subsection{Results and Observations in Standard Model}

Since the dataset is selected randomly, the distribution of duplicated samples from CIFAR-10 does not favor any particular class. As previously discussed in Section~\ref{Balanced Duplication}, this theoretically may lead to improved classification accuracy. However, the actual impact will be examined here.

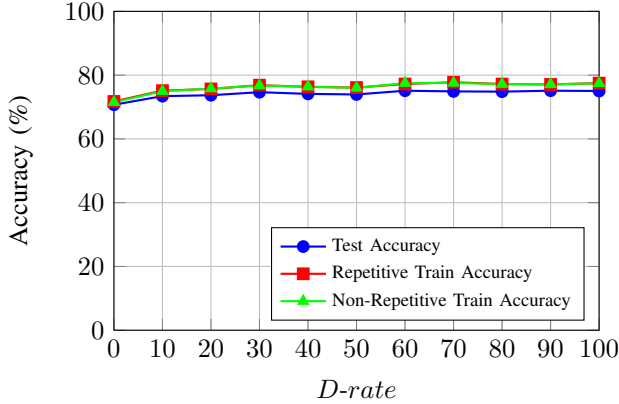
\begin{figure}[t!]
  \centering
\begin{tikzpicture}
  \begin{axis}[
      width=8cm,
      height=5.8cm,
      xlabel={$D\text{-}rate$},
      ylabel={Accuracy (\%)},
      grid=both,
      xmin=0, xmax=100,
      ymin=0, ymax=100,
      xtick={0,10,20,30,40,50,60,70,80,90,100},
      legend pos=south east,
      legend cell align={left},
  ]
    \addplot[
      color=blue,
      mark=*,
      thick,
    ] coordinates {
      (0,70.72) (10,73.41) (20,73.70) (30,74.65) (40,74.12)
      (50,73.94) (60,75.11) (70,74.90) (80,74.82) (90,75.12) (100,75.02)
    };
    \addlegendentry{\scriptsize Test Accuracy}
    
    \addplot[
      color=red,
      mark=square*,
      thick,
    ] coordinates {
      (0,71.75) (10,75.14) (20,75.67) (30,76.85) (40,76.33)
      (50,76.10) (60,77.22) (70,77.77) (80,77.21) (90,77.08) (100,77.49)
    };
    \addlegendentry{\scriptsize Repetitive Train Accuracy}
    
    \addplot[
      color=green,
      mark=triangle*,
      thick,
    ] coordinates {
      (0,71.51) (10,74.95) (20,75.79) (30,76.71) (40,76.37)
      (50,76.00) (60,77.49) (70,77.55) (80,77.18) (90,77.11) (100,77.42)
    };
    \addlegendentry{\scriptsize Non-Repetitive Train Accuracy}
  \end{axis}
\end{tikzpicture}
  \caption{The effect of adding repetitive samples to the training set on test accuracy, the accuracy of the model on the training set including duplicated samples (repetitive accuracy), and the accuracy of the model on the training set excluding duplicated samples (non-repetitive accuracy).}
  \label{fig:duplication_line_chart}
\end{figure}

Our key findings are summarized in Figure~\ref{fig:duplication_line_chart}, which shows the test and training accuracy for models trained with varying levels of data duplication. The experiments reveal that without duplication, the test accuracy is 70.72\%. With 30\% duplication (\(D\)-rate = 30), the test accuracy improves to 74.12\%, reflecting a modest 3\% increase. This suggests that increasing the number of duplications, which adds more data for training and reduces efficiency, does not lead to a substantial improvement in accuracy. Increasing the duplication rate from 30\% to 60\% results in only small improvements, increasing test accuracy from 74.12\% to 75.11\%. However, beyond this point, additional duplication yields diminishing returns, and in some cases, even slight performance degradation. This shows that excessive duplication may introduce redundancy without contributing significant diversity to the training process.

\subsection{Results and Observations in Robust Model}

Adversarially trained models exhibit a more complex response to data duplication, where repetition does not improve robustness. Unlike standard models, which benefit from uniform repetition across classes, adversarially trained models are sensitive even to uniform duplication. Such repetition undermines adversarially trained models' accuracy.

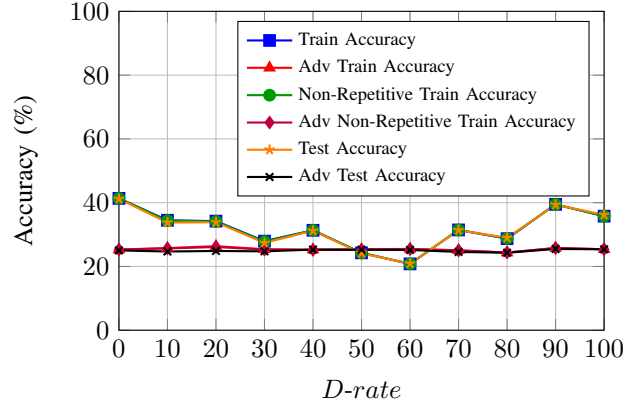
\begin{figure}[t!]
  \centering
\begin{tikzpicture}
  \begin{axis}[
      width=8cm,
      height=5.8cm,
      xlabel={$D\text{-}rate$},
      ylabel={Accuracy (\%)},
      grid=both,
      xmin=0, xmax=100,
      ymin=0, ymax=100,
      xtick={0,10,20,30,40,50,60,70,80,90,100},
      legend pos=north east,
      legend cell align={left},
  ]
    \addplot[
      color=blue,
      mark=square*,
      thick,
    ] coordinates {
      (0,41.35) (10,34.48) (20,34.22) (30,27.94) (40,31.34) 
      (50,24.27) (60,20.81) (70,31.46) (80,28.72) (90,39.47) (100,35.77)
    };
    \addlegendentry{\scriptsize Train Accuracy}

    \addplot[
      color=red,
      mark=triangle*,
      thick,
    ] coordinates {
      (0,25.30) (10,25.77) (20,26.36) (30,25.40) (40,25.27) 
      (50,25.39) (60,25.50) (70,25.08) (80,24.33) (90,25.80) (100,25.39)
    };
    \addlegendentry{\scriptsize Adv Train Accuracy}

    \addplot[
      color=green!60!black,
      mark=*,
      thick,
    ] coordinates {
      (0,41.35) (10,34.35) (20,34.09) (30,27.81) (40,31.33) 
      (50,24.25) (60,20.82) (70,31.42) (80,28.78) (90,39.47) (100,35.77)
    };
    \addlegendentry{\scriptsize Non-Repetitive Train Accuracy}

    \addplot[
      color=purple,
      mark=diamond*,
      thick,
    ] coordinates {
      (0,25.30) (10,25.65) (20,26.10) (30,25.29) (40,25.17) 
      (50,25.20) (60,25.41) (70,25.02) (80,24.37) (90,25.77) (100,25.39)
    };
    \addlegendentry{\scriptsize Adv Non-Repetitive Train Accuracy}

    \addplot[
      color=orange,
      mark=star,
      thick,
    ] coordinates {
      (0,41.16) (10,33.76) (20,33.87) (30,27.39) (40,31.19) 
      (50,24.10) (60,20.90) (70,31.38) (80,29.00) (90,39.37) (100,36.21)
    };
    \addlegendentry{\scriptsize Test Accuracy}

    \addplot[
      color=black,
      mark=x,
      thick,
    ] coordinates {
      (0,25.03) (10,24.69) (20,24.86) (30,24.74) (40,25.29) 
      (50,25.24) (60,25.18) (70,24.55) (80,24.30) (90,25.49) (100,25.38)
    };
    \addlegendentry{\scriptsize Adv Test Accuracy}

  \end{axis}
\end{tikzpicture}
  \caption{Impact of repetitive data in the training set on training accuracy, test accuracy, and the robustness (adversarial accuracy) of an adversarially trained model.}
  \label{fig:repetition_effect}
\end{figure}

This is illustrated in Figure~\ref{fig:repetition_effect}, which shows the impact of duplication on both accuracy (i.e., the ability to correctly classify clean data) and adversarial accuracy (i.e., robustness—the ability to correctly classify data with adversarial perturbations) across different levels of uniform repetition. Without duplication, test accuracy is 41.16\%. At 30\% duplication, accuracy decreases to 27.39\%, and at 60\% duplication, it further drops to only 20.90\%. This suggests a significant negative effect of duplication on accuracy, without improving adversarial accuracy (or robustness). Furthermore, this supports our claim in Proposition~\ref{proposition:2} that repetitive data negatively impacts accuracy.

This suggests that adding duplications in adversarially trained models does not lead to a general improvement in robustness or accuracy, and in fact, it may hinder the model's ability to classify clean and adversarial data correctly.

\section{Conclusion and Future Work}

This paper investigates the impact of duplicated samples in image classifier DNNs. First, a theoretical analysis explores how duplication influences generalization. Then, experimental studies are conducted. Our findings suggest that while duplication can sometimes aid in refining decision boundaries, it does not always improve generalization. In particular, under non-uniform duplication or adversarial training conditions, duplicated data may negatively impact generalization. The observations from our experiments conclude that duplication not only negatively impacts efficiency—since more data must be trained—but also requires careful handling, as it can harm the model's generalization ability depending on the distribution and nature of the repetitive data.

Our results are based on selecting duplications from a Gaussian distribution and training a DNN on CIFAR-10 as a toy example. Future work can explore the impact of duplication in more practical scenarios, such as FL, where duplication is more likely due to clients lacking knowledge of each other's datasets. Another research direction involves developing methods for private deduplication. Additionally, duplication may have privacy implications, as duplicated samples could be more easily revealed. However, the actual effect on privacy requires further investigation.


   \section*{Acknowledgment}
This work was supported by the Flemish Government through the Cybersecurity Research Program with grant number: VOEWICS02 and in part by the Research Council KU Leuven projects IF/C1 From Website Fingerprinting to App Fingerprinting: Inferring private user activity from encrypted network traffic, and the AIDE project funded by the Belgian SPF BOSA under the programme “Financing of projects for the development of artificial intelligence in Belgium” with reference number 06.40.32.33.00.10." Sumanta Sarkar was supported by UKRI through the EPSRC grants EP/T014784/1 and EP/X036669/1.





%

\bibliographystyle{IEEEtran}
\bibliography{references}

\appendices
\section{Theoretical Model: Gaussian Data Generation and SVM Classification}\label{sec::Theoretical-Model}

The data is generated using two Gaussian distributions with mean vectors $\boldsymbol{\mu}_1$ and $\boldsymbol{\mu}_2$, and a shared covariance matrix $\Sigma$. The probability density function for class $\mathcal{C}_k$ where $k \in \{1, 2\}$ is given by:
\begin{equation}
p(\mathbf{x}|\mathcal{C}_k) = \frac{1}{(2\pi)^{d/2}|\Sigma|^{1/2}} \exp \left( -\frac{1}{2}(\mathbf{x} - \boldsymbol{\mu}_k)^\top \Sigma^{-1} (\mathbf{x} - \boldsymbol{\mu}_k) \right),
\end{equation}
where $\mathbf{x} \in \mathbb{R}^d$ is a data point in $d$-dimensional space.

In the experiments, an SVM with a Radial Basis Function (RBF) kernel is used to compute the decision boundaries. The RBF kernel is defined as:
\begin{equation}
K(\mathbf{x}_i, \mathbf{x}_j) = \exp \left( -\gamma |\mathbf{x}_i - \mathbf{x}_j|^2 \right),
\end{equation}
where $\gamma > 0$ is a parameter that controls the width of the Gaussian function.

The SVM solves the following optimization problem to find the optimal separating hyperplane:
\begin{equation}
\min_{\mathbf{w}, b, \xi} \frac{1}{2} |\mathbf{w}|^2 + C \sum_{i=1}^{n} \xi_i,
\end{equation}
subject to the constraints:
\begin{equation}
y_i (\mathbf{w}^\top \phi(\mathbf{x}_i) + b) \geq 1 - \xi_i, \quad \xi_i \geq 0,
\end{equation}
where $\mathbf{w}$ is the weight vector, $b$ is the bias term, $\xi_i$ are the slack variables, $C$ is the regularization parameter, and $\phi(\mathbf{x})$ is the feature transformation induced by the RBF kernel.

\end{document}